\RequirePackage{fix-cm}
\documentclass[smallcondensed]{svjour3}     
\smartqed  
\usepackage{graphicx}
\usepackage{algpseudocode}
\usepackage{amsmath}
\usepackage{algorithm}
\usepackage{algorithmicx} 
\usepackage{algpseudocode}
\usepackage{threeparttable}
\usepackage{color}

\begin{document}

\title{Hybrid Multisource Feature Fusion for the Text Clustering
}
\subtitle{}


\author{Jiaxuan Chen         \and
        Shenglin Gui 
}


\institute{Jiaxuan Chen \at
              University of Electronic Science and Technology of China \\
              \email{chenjiaxuan@std.uestc.edu.cn}           
           \and
           Shenglin Gui \at
              University of Electronic Science and Technology of China\\
              \email{shenglin\_gui@uestc.edu.cn}
}

\date{Received: date / Accepted: date}

\maketitle

\begin{abstract}
The text clustering technique is an unsupervised text mining method which are used to partition a huge amount of text documents into groups. It has been reported that text clustering algorithms are hard to achieve better performance than supervised methods and their clustering performance is highly dependent on the picked text features. Currently, there are many different types of text feature generation algorithms, each of which extracts text features from some specific aspects, such as VSM and distributed word embedding, thus seeking a new way of obtaining features as complete as possible from the corpus is the key to enhance the clustering effects. In this paper, we present a hybrid multisource feature fusion (HMFF) framework comprising three components, feature representation of multimodel, mutual similarity matrices and feature fusion, in which we construct mutual similarity matrices for each feature source and fuse discriminative features from mutual similarity matrices by reducing dimensionality to generate HMFF features, then k-means clustering algorithm could be configured to partition input samples into groups. The experimental tests show our HMFF framework outperforms other recently published algorithms on 7 of 11 public benchmark datasets and has the leading performance on the rest 4 benchmark datasets as well. At last, we compare HMFF framework with those competitors on a COVID-19 dataset from the wild with the unknown cluster count, which shows the clusters generated by HMFF framework partition those similar samples much closer. 

\end{abstract}

\section{Introduction}
\label{intro}
The information in people's daily lives has shown an explosive growth nowadays due to the rapid development of the Internet. For these large-scale and disorderly text datasets, how to dig out the most valuable information has always been a key research topic in the field of natural language processing \cite{Ref1}. Clustering are sort of unsupervised learning method, which can divide a large amount of unknown text information into a set of intelligible clusters. By finding such distribution information from corpus, the search range can be reduced on a large scale and target information can be easily found though filtering.

The most widely used text model in the early days is the Vector Space Model (VSM) \cite{Ref4}, where each document is represented as a list of weighted terms, based on the frequency-inverse document frequency(TF-IDF) \cite{Ref2}. However, VSM given TF-IDF only can exploit the advantages of text information from the word level, it ignores the potential semantics in the text. Thus result in reducing the value of the information. To address these problems, topic modeling methods are incorporated.L. M. Aiello et al.\cite{Ref3} investigates several popular topic modeling methods, including Latent Dirichlet Allocation (LDA), mallet LDA, Hierarchical Dirichlet Process (HDP), etc. LDA model considers the potential topic of the document to cluster the text, however, this technology is still based on the discrete representation of the text and does not consider the context. Thus, it ignores the location information of the word. Since word location is a kind of key information in the text and the semantic will be very different if the location of the word is different. Mikolov et al. proposed Word2vec\cite{Ref5} and Doc2vec\cite{Ref15}, which are two types of distributed word embeddings (DWE). DWE takes the context of words into account and learns the semantic and grammatical information. However, since word and vector are one-to-one relationship in DWE model, it is difficult to optimize the algorithm itself in the face of some text with unsatisfactory processing effect.

In recent years, more optimization text feature selection algorithms has been proposed to improve clustering performance. Hujino et al. \cite{Ref6} utilized Jensen-Shannon divergence to measure documents similarity. Laith et al. \cite{Ref2} proposed Multi-objectives-based text clustering algorithm (MKM) to improve VSM performance. However, these two methods respectively use just one type of text represents algorithm to select features from documents corpus, both of which de facto loss some critical and useful text features. Hu et al.\cite{Ref1} proposed a hybrid method which integrate VSM and LDA model, this model combine the advantage of these two models by hybrid the weighted similarity matrices between documents corpus. However, bunches of experiments need to be conducted to obtain appropriate weight parameters, which consumes large amounts of time. Hassani et al. \cite{Ref8} proposed a text feature extract technique based on Nonnegative Matrix Factorization and Latent Semantic Analysis (NMF-FR). NMF-FR employ NMF method to reduce dimensionality of term-document matrix which is extracted by Latent Semantic Analysis (LSA) model, then agglomerate term vectors to create a new feature space. This method take the dimensionality of feature vectors into account and improve the performance of LSA text represent algorithm. However, NMF-FR ignore the potential semantics in the text, thus result in text information loss and affect agglomeration performance. Hybrid harmony search algorithm (FSHSTC)\cite{Ref7} and hybrid particle swarm optimization algorithm (H-FSPSOTC) \cite{Ref9}, etc. also widely used in text clustering area. FSHSTC based on harmony search algorithm, which displace uninformative options with a replacement set of fine options to improve feature selection performance. H-FSPSOTC apply genetic operators to improve particle swarm optimization (PSO) algorithm performance. Nevertheless, the performance of these algorithms fluctuate wildly in different benchmark datasets. The solution this paper proposed, in contrast, keep leading performance in each experiment.

In this paper, a Hybrid Multisource Feature Fusion (HMFF) framework is proposed by fusing two kinds of feature extraction algorithms based on discrete representation and distributed representation. HMFF analyzes document features from multiple perspectives to ensure the stability and accuracy of the algorithm when processing different kinds of documents and effectively avoid the failure of the algorithm. Below we summarize the contributions of this paper:
\begin{itemize}

  \item [1)] We proposed a new text feature fusion framework, HMFF, which integrate with three different text representation source to extract more comprehensive features from input samples. The benchmark experiments results show HMFF performs overall better than other competitors.
  \item [2)] The previous text clustering algorithms usually compared their performance on dataset with known cluster count, however, most of them did not show how to be applied on dataset in the wild with unknown cluster count. It is a huge challenge for text representation algorithms to obtain good clustering effect on a stochastic wild dataset with unknown cluster count. In this paper, we show how to apply HMFF framework on COVID-19 public opinion dataset from Twitter apart from standard datasets. Our HMFF shows a better performance than other algorithms under various cluster number settings.
  \item [3)] HMFF framework has good scalability, hence it could easily integrate with additional feature sources into HMFF feature matrix from other text representation algorithms to enlarge feature coverage of corpus without significantly increasing overall feature-computing time.
  \end{itemize}


\section{Preliminaries}
\label{sec:2}
In this section, we briefly show the basic information which will be used in this paper, including text preprocessing technique, vector space model, topic model, word embedding and K-means clustering. 
\subsection{Text preprocessing}
\label{sec:2.1}

 Text preprocessing is a key work in text clustering, document retrieval or other technologies related to natural language processing. The purpose of text preprocessing is to transform the original text into a format that the program can understand and handle easily. The preprocessing of English documents usually consists of POS tagging, stemming and stop words removal.
  
  \begin{itemize}
    \item [1)]
    \begin{itemize}
      \item [i)]
    Tagging words with POS tags according to their meanings and context content. Effective POS tagging can help to improve the efficiency of tokenization. 
      \item [ii)]
    Splitting a text document into tokens and deleting empty statements and documents, in which each word or symbol is called a token. 
    \end{itemize}
      
    \item [2)]
    Stemming is the process of reducing word by removing the affix to get the most common way to write words, namely, the roots. The Porter stemmer is the common stemming method embraced in content mining \cite{Ref11}. 
    
    
    \item [3)]
Removing high frequency words, such as he, she, I, etc. Because these words appear in almost every text. They don't bring symbolic features to the document. In addition, this kind of words will also affect the feature extraction of the document, resulting in the accuracy of the subsequent document clustering operation is reduced. The list of forbidden words can be found in this website$\footnote{http://members.unine.ch/jacques.savoy/clef/englishST.txt}$.
  \end{itemize}

\subsection{Vector Space Model}
\label{sec:2.2}

In the late 1960s, Salton and others first proposed the vector space model (VSM). VSM is a common model that easy to implement which describe documents as a series of vectors composed of keywords. In this process, each text is abstracted into a vector. The length of the vector equals to the number of document feature words, and the value is the weight of the feature value. VSM represents documents as follows:
\begin{equation}
V(d) = {w_1(d),w_2(d),...,w_n(d)}
\end{equation}
where $w_i(d)$ is the TF-IDF weight of dictionary token $i$ in document $d$.

TF-IDF\cite{Ref12} is a common weighting technology in information retrieving and data mining. The method is based on statistics and is used to calculate the importance of words in corpus. Compared with one-hot method, TF-IDF can extract informative words from common but unimportant words. Let $n_i(d)$ be the count of occurrences of each term in each text, the following equation is used to expressed Term and Inverse Document Frequencies.
\begin{equation}
w_i(d)=TFIDF_i(d)=\frac{n_i(d)}{t(d)}\cdot \left(\frac{\log_n}{df_{i}}\right)
\end{equation}
where $n$ is the number of all texts in whole dataset, $t(d)$ is the number of all tokens in text $d$ and $df_i$ means how many text contains the token $i$.

\subsection{Topic model}
\label{sec:2.3}
 Topic model is a kind of statical model for using the abstract “topics” to represent documents. Latent Dirichlet Allocation is a classical topic model that was proposed by BLEI in 2002\cite{Ref13}. It is a generative model based on probability. In the topic model, the text is composed of selected words from certain topics, and the specific words in the text can reflect the specific topics. Therefore, in Latent Dirichlet Allocation model, each topic is considered as the probability distribution of keywords, and each document is represented as the probability distribution of several topics in the dataset\cite{Ref1}.

Fig.1 shows the Graphical model representation of LDA. The figure utilizes boxes to represent replicates, i.e. repeated entities. The outer plate represents documents, while the inner plate represents the repeated word positions in a given document, and each position is associated with a choice of topic and word\cite{Ref14}.  In the figure, only $w$ is greyed, because it is the only observable variable, and the others are all latent variables.

\begin{figure}[htb]
  \includegraphics[width=0.5\textwidth]{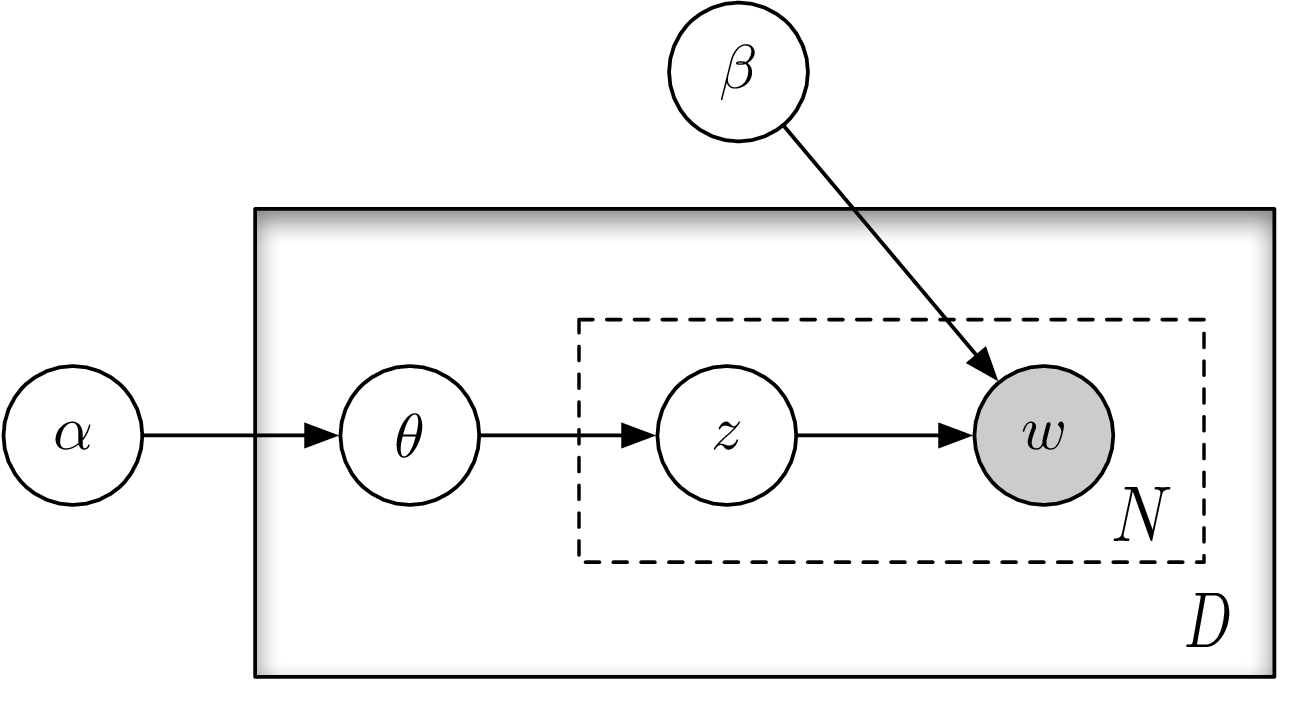}
  \centering
\caption{Graphical model representation of LDA}
\label{fig:1}
\end{figure}

Suppose there are $D$ documents, and each of them has $N_i$ words, where $i\in \{1,...,M\}$. When generating the whole corpus, LDA assumes the following steps:
\begin{itemize}
  \item [1)] 
  Choose $\varphi_k\sim Dirichlet(\beta)$, which is the word distribution of the topic $k \in \{1,...,K\}$.
  \item [2)]
  For each document $d_i$, $i \in \{1,...D\}$:
  \begin{itemize}
    \item [i)] 
    Choose $\theta_i\sim Dirichlet(\alpha)$, which is the topic distribution of the document.
    \item [ii)] 
    For each word $w_{ij}$,$j\in \{1,...,N_i\}$:
    \begin{itemize}
      \item [a)]
      Choose $z_{ij} \sim Multinomial(\theta_i)$, which is the topic of word $w_{ij}$.
      \item [b)]
      Choose $w_{ij} \sim Multinomial(\varphi_{z_{ij}})$ 
    \end{itemize}
  \end{itemize}
\end{itemize}
There are two major approaches to learn an LDA model, which are variational inference and Gibbs sampling. The learning process of LDA ends up with a set of statistical parameters, including $\alpha,\beta,\theta$ and $\varphi$, that maximize the log likelihood of the corpus. In other words, the learning algorithm maximizes the following probability:
\begin{equation}
P(W, Z, \theta, \varphi ; \alpha, \beta)=\prod_{i=1}^{K} P\left(\varphi_{i} ; \beta\right) \prod_{j=1}^{M} P\left(\theta_{j} ; \alpha\right) \prod_{t=1}^{N} P\left(Z_{j t} \mid \theta_{j}\right) P\left(W_{j t} \mid \varphi_{Z_{j t}}\right)
\end{equation}

\subsection{Distributed word embeddings}
\label{sec:2.4}
Distributed Word Embedding\cite{Ref15} is an approach of representing words in a corpus by mapping them into an n-dimensional vector space, in which words with similar meaning are expected to lie close to each other. Word2vec and Doc2Vec are two typical distributed word embedding algorithms, which are suitable for data in different formats.
\begin{figure}[htb]
\centering
  \includegraphics[width=1\textwidth]{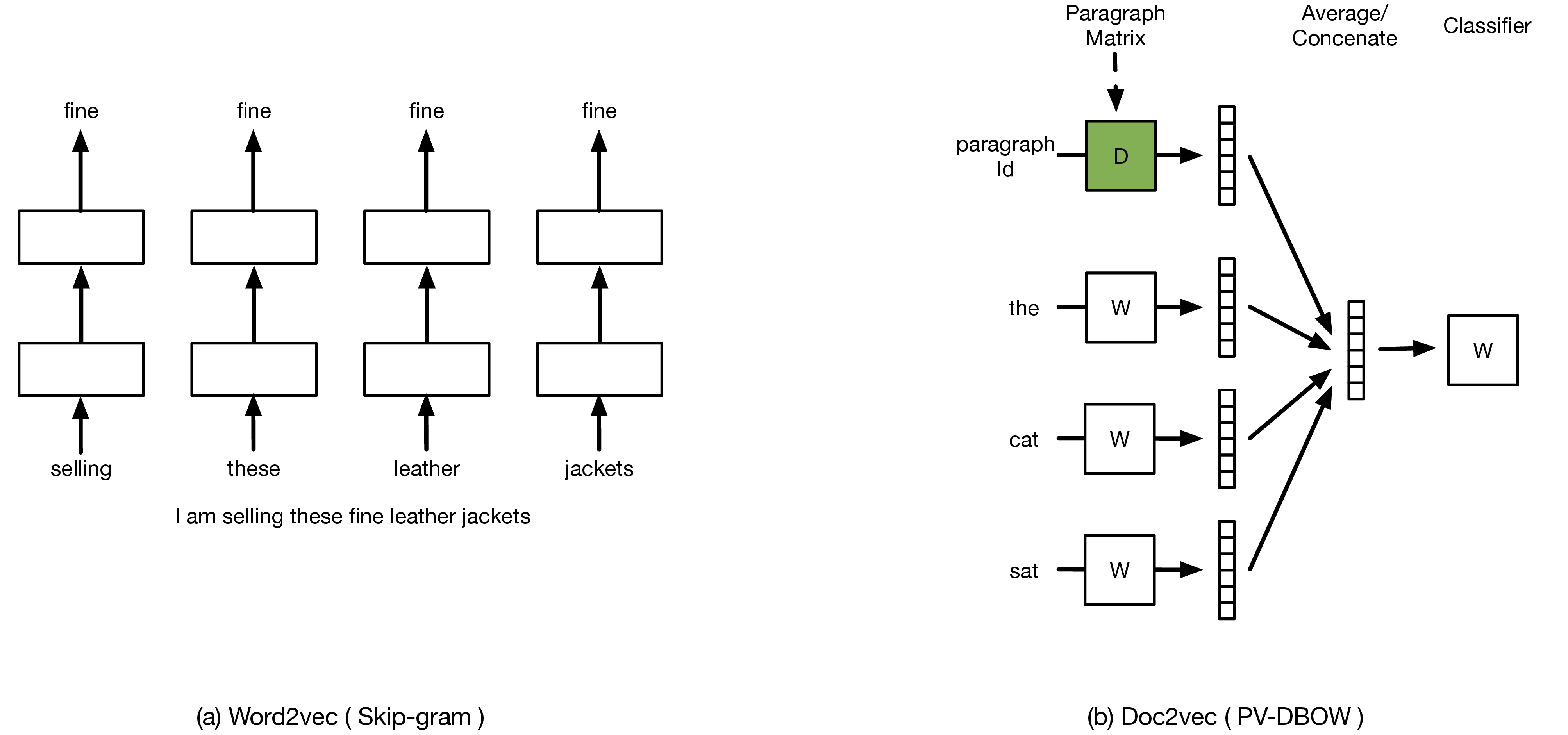}
\caption{two typical word embedding framework}
\label{fig:3}
\end{figure}
\subsubsection{Word2vec algorithm}
\label{sec:2.4.1}
Word2vec is a word vectorization method based on two-layer neural network. Given enough training data, word2vec can fit all the words into an $n$-dimensional space automatically. There are two major strategies in doing so, continuous bag of words (CBOW) and Skip-gram. Skip-gram strategy predicts a word’s meaning using a random close-by word, while CBOW takes into account of the whole context of the word defined by a window size\cite{Ref16}. In practice, the Skip-gram model has proved to have better overall performance.



\subsubsection{Doc2vec algorithm}
\label{sec:2.4.2}
Doc2vec paragraph vector method\cite{Ref17} is an unsupervised algorithm, which can learn fixed length feature representation from variable length text (such as sentences, paragraphs and documents). The algorithm trains the predicted words in the document to use a single dense vector to represent each document. It is inspired by the word vector learning, paragraph vector can predict the next word according to the given context samples from the paragraph. The original paper mentioned two methods: distributed memory model of paragraph vectors, PV-DM and distributed bag of words version of paragraph vectors, PV-DBOW\cite{Ref18}.

When compared with bag of words model, Doc2vec has some unique advantages. Firstly, the feature vectors can inherit the semantic information of words. Secondly, word order is considered like the N-gram model on the basis of small context, which retains more information about paragraphs, including word order. For the data set in word bag format, due to the context information of words cannot be obtained, Word2vec is used as a part of feature extraction. Thus, for the data sets that can express word order, Doc2vec technology incline to have better performance.

\subsection{K-means text clustering algorithm}
\label{sec:2.5}
K-means algorithm\cite{Ref19} is a basic partition method in clustering analysis. Firstly, the samples are roughly clustered, and then the clustering results are continuously optimized according to some correction principle until the clustering results are reasonable. Each data points inside the dataset is a vector of document features. K-means algorithm mainly consists of the following four steps:
\begin{itemize}
  \item [1.]
  From $n$ texts $x = x_1,x_2,...,x_n$ to be clustered, $k$ texts are randomly selected as the initial clustering center $a = a_1,a_2,...,a_k$.
  \item [2.]
  For each sample $x_i$ in the data set, calculate its distance to the each cluster center and assign it to the cluster corresponding to the cluster center with the shortest distance.
  \item [3.]
  For each category $a_j$, recalculate its cluster centers $a_{j}=\frac{1}{\left|c_{i}\right|} \sum_{x \in c_{i}} x$.
  \item [4.]
  Repeat step2 and step3 until the specified number of iterations $Step_{max}$ is reached or the termination condition $\left|E_{n+1}-E_n\right|\leq \varepsilon$ is satisfied,where the standard measure function $E=\sum_{i=1}^{k} \sum_{x \in C_{i}}\left|x-\bar{x}_{i}\right|^{2}$, $\bar{x}_{i}$ is the text in the center of $C_{i}$.
  
\end{itemize}

\section{Proposed method}
\label{sec:3}
In this section, we present the details about proposed text feature fusion framework, Hybrid Multisource Feature Fusion (HMFF) framework, which consists of three main parts: feature representation of multi-model, mutual similarity matrices construction, dimension reduction and features fusion. Notice that in this section, the whole document dataset is already assumed to be preprocessed and be tokenized into token-document dataset $D_{token}$. 

As HMFF framework shown in Fig.3, we construct mutual similarity matrices for each feature source and fuse discriminative features from mutual similarity matrices by reducing dimensionality to generate HMFF features $D_{HMFF(n \times h)}$, , where $n$ is the number of documents and $h$ is the dimension of hybrid features. In order to detect the performance of HMFF framework, k-means clustering algorithm is configured to partition $D_{HMFF}$ into groups. Accuracy, F-measure and Silhouette Coefficient are used as clustering judgements to measure the algorithm compared with other similar algorithms.

\begin{figure}[htb]
  \includegraphics[width=1\textwidth]{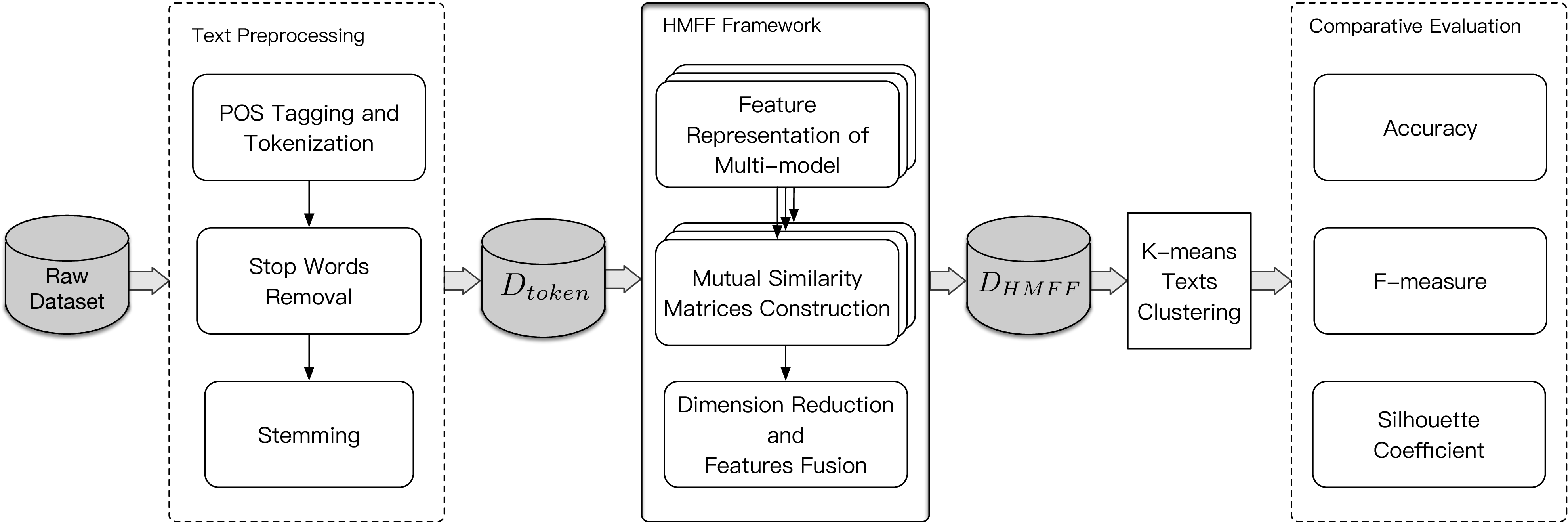}
\caption{flowchart of applying the proposed HMFF framework.  The left dotted rectangle shows main text preprocessing steps that convert raw datasets to $D_{token}$. The main parts of HMFF framework are presented at middle rectanglen, which generate HMFF feature-document matrix $D_{HMFF}$ from token-document dataset $D_{token}$. And comparative evaluation indexes are shown in the right dotted rectangle to measure HMFF framework performance.}
\label{fig:3}
\end{figure}

\subsection{Features representation of multi-model}
\label{sec:3.1}
In this selection we will introduce the mathematical representation of document features under VSM $m_{vsm}$, topic model $m_{topic}$ and distributed word embedding $m_{emb}$. By using text representation method $m_i$, we convert the representation of dataset from a token-document format $D_{token}$ to a feature-document matrix $Mat_{feat}(m_i)$ as follows:
\begin{equation}
Mat_{feat}(m_i)=[Vec(d_1,m_i)^T,…,Vec(d_{n-1},m_i)^T,Vec(d_n,m_i)^T]
\end{equation}
where $m_i \in \{m_{vsm}, m_{topic}, m_{emb}\}$, $n$ is number of document in the dataset, $Vec(d_j,m_i)^T$ is the vector that represent document $d_j$ under model $m_i$.

\subsubsection{Features from Vector Space Model}
\label{sec:3.1.1}
Vector Space Model (VSM) use TF-IDF weight as features to represent documents, and all documents using one dictionary which include every tokens from the dataset. In order to ensure the consistency of the document feature vector length, we assign the weight of the tokens in the dictionary that does not appear in document $d_i$ to zero. Hence, $d_i$ can be represented in an $N_{vsm}$-dimensional vector space, where $N_{vsm}$ is total number of unique tokens from the whole dataset. Let $n$ be the number of documents in a collection, document $d_i$ under VSM ${m_{vsm}}$ is represented as follows:
\begin{equation}
Vec(d_i,m_{vsm})=\left(w(t_1,d_i),w(t_2,d_i),...,w(t_{N_{vsm}},d_i)\right)
\end{equation}
where $ i \in \{1,2,...,n\}$,$1 \leq j \leq N_{vsm} $ and $w(t_j,d_i)$ represents the weight of token $t_j$ in document $d_i$, it can be measured under TF-IDF encoding or one hot encoding.

\subsubsection{Features from topic model}
\label{sec:3.1.2}
Documents can be represented as probability distribution of each topic under topic model, the number of topic determines the length of the feature vector. In this paper we use topic coherence measurement to choose optimal topic numbers. 

Coherence score\cite{Ref34} was proposed by Michael in 2015, this evaluation model is more reasonable than the traditional log perplexity measurement of topic model. We use $C_{UMass}$\cite{Ref34} as  coherence score, which based on document cooccurrence counts, to evaluate whether one topic model is good or not.
\begin{equation}
C_{U M a s s}=\frac{2}{N *(N-1)} \sum_{i=2}^{N} \sum_{j=1}^{i-1} \log \frac{P\left(w_{i}, w_{j}\right)}{P\left(w_{j}\right)}
\end{equation}
where $w_i$ and $w_j$ are two different words in the dataset, $P(w_i,w_j)$ is the proportion of documents that contain both words $w_i$ and $w_j$ to all documents in the dataset, $p(w_j)$ is the proportion of documents that contain words $w_j$ to all documents in the dataset, $N$ is the top words number which are chosen from a topic. 

In this formula, $P(w_i,w_j)$ and $P(w_j)$ are determined by topic number and other parameters are all fixed when a dataset and topic model are given. A higher score means that the topic has better interpretability in semantics, and the semantics within the topic are more coherent. Then we can use this formula to calculate an optimal topic number $N_{topic}$.
\begin{equation}
N_{topic} = argmax\{C_{UMass}\}
\end{equation}
Let $n$ be the number of documents in a collection, document $d_i$ under Topic model $m_{topic}$ can be represented in the following:
\begin{equation}
Vec(d_i,m_{topic})=\left(w(t_1,d_i),w(t_2,d_i),...,w(t_{N_{topic}},d_i)\right)
\end{equation}
where $i \in \{1,2,...,n\}, 1\leq j \leq N_{topic}$ and $w(t_j,d_i)$ represents the probability distribution of topic $t_j$ of document $d_i$, $n$ is the number of documents, $t_{topic}$ is topic num which equals to optimal topic num $N_{topic}$.

\subsubsection{Features from distributed word embedding}
\label{sec:3.1.3}
Word2vec and Doc2vec are two types of distributed word embeddings. The former is suitable for corpus in format of tokens, and if we have word order information of each document, we can use Doc2vec to extract paragraph information directly.
\begin{itemize}

  \item [i)]
In Word2vec model, each unique token $\tau_k$ in corpus is mapped to an $N_{w2v}$-dimensional numerical vector as follows:
\begin{equation}
Vec(\tau_k,m_{w2v})=\left(v(t_1,\tau_k),v(t_2,\tau_k),...,v(t_{N_{w2v}},\tau_k)\right)
\end{equation}
where $v(t_j,\tau_k)$ is the value of feature $t_j$ of the vector of token $\tau_k$, $1 \leq j \leq N_{w2v}$.  Let suppose $N(\tau,D)$ is total number of unique tokens from the whole dataset $D$, $1 \leq k \leq N(\tau,D)$. Dimension of word vector $N_{w2v}$ is determined by the number of hidden layer nodes in neural network that trained  model $m_{w2v}$. 

Then we use one ‘focus word’ to represent document $d_i$ by averaging the word vectors of all tokens in $d_i$. The representation of $d_i$ under Word2vec model $m_{w2v}$ is as follows:
\begin{equation}
\begin{aligned}
Vec(d_i,m_{w2v})&=\frac{1}{N(\tau,d_i)}\sum_{k =1}^{N(\tau,d_i)} v(t_j,\tau_k)\\&=\left(w(t_1,d_i),w(t_2,d_i),...,w(t_{N_{w2v}},d_i)\right)
\end{aligned}
\end{equation}
where $w(t_j,d_i)$ represents the $j$th feature in average word vector of document $d_i$, $N(\tau,d_i)$ is the number of tokens in $d_i$.

\item [ii)]
In Doc2vec model, text feature representation is considered from the perspective of paragraphs. Compared with Word2vec, Doc2vec can learn fixed-length feature representations from variable-length texts, text representation format under this model is shown as:

\begin{equation}
Vec(d_i,m_{d2v})=\left(w(t_1,d_i),w(t_2,d_i),...,w(t_{N_{d2v}},d_i)\right)
\end{equation}
where $i \in \{1,2,...,n\}, 1\leq j \leq N_{d2v}$, and $w(t_j,d_i)$ represents the value of features $t_j$ in vector of document $d_i$.
\end{itemize}
Finally, we give document under distributed word embedding representation (include Word2vec model and Doc2vec model) a unified format as follows:
\begin{equation}
Vec(d_i,m_{emb})=\left(w(t_1,d_i),w(t_2,d_i),...,w(t_{N_{emb}},d_i)\right), emb \in \{w2v, d2v\}
\end{equation}

\subsection{Mutual similarity matrices construction}
\label{sec:3.2}
In HMFF framework, feature fusion from different algorithms is a key step, therefore we need design a unified document feature representation format for each feature source. We construct mutual similarity matrix which shows relationship between documents to represent document for the following reasons: 1) Dimension of similarity matrix is only determined by document number $n$ in the dataset $D$. Compared with feature-document matrix $Mat_{feat}(m_i)$, in which dimension is determined by text representation algorithm's type, it makes preparation for dimension reduction and features fusion. 2) We can choose optimal similarity measures to match various text representation model, in this way more features are kept in the matrix.

  By calculating the mutual similarity between each two documents under specific distance formula in this section, we can construct one mutual similarity matrix for each type of document representation algorithm in the following:
  \begin{equation}
  {Mat_{sim}(m_i)}=\left[\begin{array}{cccc}
  1          			&\cdots & Sim(d_1,d_{n-1}) & Sim(d_1,d_n)  \\
  \vdots 			& \ddots & \vdots & \vdots  \\
  Sim(d_{n-1},d_1)	& \cdots  &\cdots& Sim(d_{n-1},d_n) \\
  Sim(d_n,d_1)  	& \cdots & Sim(d_n,d_{n-1})  & 1\\
  \end{array}\right]
  \end{equation}
  where $1\leq i,j \leq n$, $n$ is document number in a collection, and $Sim(d_i,d_j)$ is similarity score between $d_i$ and $d_j$. Note that the value range of $Sim(d_i,d_j)$ is between 0 and 1.
 
\subsubsection{Cosine-Euclidean formula}
\label{sec:3.2.1}
The Cosine similarity measure and Euclidean distance measure are commonly used in text clustering domain to calculate mutual similarity between each feature vector of documents\cite{Ref20,Ref21,Ref22}. Let $d_i$ and $d_j$ be the two documents in the dataset, the similarity is calculated under these two formulas by the following.
\begin{equation}
\operatorname{Cos}\left(d_{i}, d_{j}\right)=\frac{\sum_{k=1}^{{N_{model}}} w\left(t_k, d_{i}\right) \times w(t_k, d_{j})}{\sqrt{\sum_{k=1}^{{N_{model}}} w(t_k, d_{i})^{2}} \sqrt{\sum_{k=1}^{{N_{model}}} w(t_k, d_{j})^{2}}}
\end{equation}
\begin{equation}
Euc\left(d_{i}, d_{j}\right)=\left(\sum_{k=1}^{{N_{model}}}\left|w\left(t_k, d_{i}\right)-w\left(t_k, d_{j}\right)\right|^{2}\right)^{1 / 2}
\end{equation}
where $w\left(t_k, d_{i}\right)$ is the weight of feature $t_k$ in document $d_i$, $w\left(t_k, d_{j}\right)$ is the weight of feature $t_k$ in document $d_j$. $model\in\{vsm,emb\}$, $N_{model}$ $\sum_{k=1}^{{N_{model}}} w\left(t_k, d_{i}\right)^2$ and $\sum_{k=1}^{{N_{model}}} w\left(t_k, d_{j}\right)^2$ are the summation of all features weight square of the document $d_i$ and $d_j$ from $k =1$ to $N_{model}$ respectively. 

HMFF technique use a kind of multi-objectives method to improve the performance of similarity matrix, which combines Cosine Similarity measure[eq14] and Euclidean Distance measure[eq15], namely, Cosine-Euclidean function. Since $w\left(t_k, d_{i}\right),w\left(t_k, d_{j}\right) \ge 0$, the range of $Cos(d_i,d_j)$ is between 0 and 1. However, $0\leq Euc(d_i,d_j)< +\infty$ and small value means high mutual similarity between two documents. This paper use Min-max feature scaling to normalize Euclidean distance and convert the function to measure mutual similarity between documents but not distance.

\begin{equation}
Euc_s(d_i,d_j)=\frac{1-Euc(d_i,d_j)-\min(1-Euc(d_i,d_j))}{\max(1-Euc(d_i,d_j))-\min(1-Euc(d_i,d_j))}
\end{equation}

Consine-Euclidean formula get advantages of both measurement together hence are able to to make mutual similarity algorithm become more precision and robust. The equation is portrayed as follows.
\begin{equation}
Sim(d_i,d_j)=CE(di,dj) = \frac{Cos(d_i,d_j)+Euc_s(d_i,d_j)}{2}
\end{equation}


\subsubsection{Jensen-Shannon divergence}
\label{sec:3.2.2}
For Topic model, the documents is represented by topic probability vector of the Dirichlet distribution. Since cosine similarity and Euclidean distance are not suitable to measure similarity between two probability distributions\cite{Ref6}, we use JS(Jensen-Shannon) divergence instead of these two measurements to calculate the document similarity, so as not to lose the advantage of topic model. To measure similarity between document $d_i$ and $d_j$ in topic model, Jenssen-Shannon divergence is defined as follows:
\begin{equation}
\begin{aligned}
JS(d_i, d_j)
=\frac{1}{2}\left(\sum_{k=1}^{N_{topic}} w(t_k,d_i) \ln \frac{w(t_k,d_i)}{w(t_k,d_j)}+\sum_{k=1}^{N_{topic}} w(t_k,d_j) \ln \frac{w(t_k,d_j)}{w(t_k,d_i)}\right)
\end{aligned}
\end{equation}

where $w(t_k,d_i)$ and  $w(t_k,d_j)$ represents the probability distribution of topic $t_k$ of document $d_i$ and $d_j$ respectively. $JS(d_i,d_j)$ is in range $[0,1]$ in the expression. 0 indicates the two distributions are the same, and 1 shows that they are nowhere similar. In consideration of uniform format, we convert it to mutual similarity formula as follows.
\begin{equation}
Sim(d_i,d_j) = 1 - JS(d_i,d_j)
\end{equation}

\subsection{Dimension reduction and features fusion}
\label{sec:3.3}
The feature selection issue in HMFF framework can be defined as integrating features which are extracted from mutual similarity matrices $Mat_{sim}(m_i)$, where $m_i \in \{m_{vsm},m_{topic},m_{emb}\}$. In this process, discriminative feature directions require to be confirmed from each mutual similarity matrix. Since mutual similarity matrix is a non-negative real symmetric matrix, all its eigenvectors of different eigenvalues are orthogonal to each other. The truncated eigendecomposition of documents similarity $Mat_{sim}(m_i)$ is defined as follows.
  \begin{equation}
  Mat_{sim}(m_i) = Q^{m_i}_{(n\times n)}\Lambda^{m_i}_{(n \times n)} Q_{(n \times n)}^{m_iT}\approx Q^{m_i}_{(n\times k)}\Lambda^{m_i}_{(k \times k)} Q_{(k \times n)}^{m_iT} 
  \end{equation}
  where $Q_{m_i}$ is the $n \times k$ matrix whose $j$th columns is the orthogonal eigenvector $v_{j}$ of $Mat_{sim}(m_i)$.  $\Lambda_{m_i}$ is the diagonal matrix whose diagonal elements are the $k$ largest corresponding eigenvalues, $\lambda_{j}$, $j=1,...,k$ and $k \ll n$ (documents number in the dataset). This can be much quicker and more economical than calculating all the eigenvalues and eigenvectors in the $Mat_{sim}(m_i)$ and are able to retain most of information of $Mat_{sim}(m_i)$.
  
  Let $\xi_j$ be the normalized $k$ eigenvectors $v_j$ in the $Q^{m_i}_{(n\times k)}$, eigenvalues $\lambda_j$ be the weight of $\xi_j$.
 We extract truncated features $\lambda_j^{m_i}\xi_j^{m_i}$ from each text representation method $m_i$ and stitch them together, as a complete fused feature matrix to represent the whole dataset. This is called HMFF feature matrix in this paper. And we describe the overall procedure of HMFF framework in Algorithm 1.
  \begin{equation}
\begin{aligned}
  D_{HMFF}=(\lambda_1^{m_1}\xi_1^{m_1},\lambda_2^{m_1}\xi_2^{m_1},...,\lambda_k^{m_1}\xi_k^{m_1},\lambda_1^{m_2}\xi_1^{m_2},\lambda_2^{m_2}\xi_2^{m_2}...,\lambda_k^{m_2}\xi_k^{m_2},...)\\
{\rm where}\ m_i \in \{m(vsm),m(topic),m(emb)\}
\end{aligned}
\end{equation}

\renewcommand{\algorithmicrequire}{\textbf{Input:}}  
\renewcommand{\algorithmicensure}{\textbf{Output:}} 
\begin{algorithm}[H]
    \caption{HMFF framework}
    \label{alg:HMFF Technique}
    \begin{algorithmic}[1]
      \Require all tokens of dataset $D_{token}$
      \Ensure	HMFF features of dataset $D_{HMFF}$
      \State determine text representation models $M, m_i \in M $ based on format of document $D_{token}$.
      \For {$m_i \in M$}
      \For {$d_i \in D_{token}$}
      \State $Vec(d_i,m_i)\gets$ represent $d_i$ into a numerical representation under model $m_i$ according to [Eq5-Eq12]
      \EndFor
      \State  $Mat_{feat}(m_i)\gets$ represent whole dataset according to [Eq4]
      \State $Mat_{sim}(m_i)\gets$ build similarity matrix according to [Eq13-Eq19]
      \If {$m_i \in \{m(vsm),m(emb)\}$}
      \Return $Sim(d_i,d_j)= CE(d_i,d_j)$
      \Else \
      \Return $Sim(d_i,d_j)=1-JS(d_i,d_j)$ where $d_i, d_j$ are two different documents in the dataset.
      \EndIf
      \State $Q_{(n\times k)}^{m_i}\Lambda_{(k \times k)}^{m_i} Q_{(k \times n)}^{m_iT}\gets$ calculate truncated eigendecomposition with $k$ largest eigenvalues of  $Mat_{sim}(m_i)$
      \EndFor
      \State $D_{HMFF}\gets$ integrate all truncated features $\lambda^{m_i}\xi^{m_i}$ from different model $m_i$ according to [Eq21]
    \end{algorithmic}
\end{algorithm}

\section{Experiment details}
\label{sec:4}
We have programmed the HMFF technique using Jupyter Notebook (Python 3.8.5 64-bit) to perform the document clustering mechanism. In this section we illustrate the evaluation measures, then we explain experiment details on benchmark datasets. At last we utilize HMFF technique to conduct extended experiment about COVID-19 twitter dataset.

\subsection{Evaluation measures}
\label{sec:4.1}
Since benchmark datasets has been labeled with categories, the comparing assessments were conducted utilizing one internal evaluation measure, silhouette coefficient, and two external evaluation measures, accuracy(Ac) and F-measure(F). These measures are the common evaluation criteria used in the domain of the text clustering to evaluate the clusters accuracy\cite{Ref23}.
\subsubsection{internal evaluation measure}
\label{sec:4.1.1}
Silhouette Coefficient is a typical internal evaluation method of clustering effect. It combines the two factors of cohesion and separation, which is used to evaluate the impact of different algorithms or different operating modes of algorithms on the clustering results on the basis of the same original data\cite{Ref24}.

For text record point $i \in C_i$, the definition of Cohesion is:
\begin{equation}
a(i)=\frac{1}{\left|C_{i}\right|-1} \sum_{j \in C_{i}, i \neq j} d(i, j)
\end{equation}
where $d(i,j)$ is the distance between text points $i$ and $j$ in the cluster $C_i$.

And the Separation of text data point $i \in C_i$ is:
\begin{equation}
b(i)=\min _{k \neq i} \frac{1}{\left|C_{k}\right|} \sum_{j \in C_{k}} d(i, j)
\end{equation}

Considering the coefficient and separation, a silhouette value of one text data point $i$ is defined as below:
\begin{equation}
s(i)=\left\{
\begin{array}{c}
  \frac{b(i)-a(i)}{\max \{a(i), b(i)\}} , \left|C_{i}\right|>1\\
  0, \left|C_{i}\right|=1
\end{array}
\right.
\end{equation}
The value of the silhouette coefficient of the clustering result is between [-1,1]. The larger the value, the closer the similar samples are. The farther the different samples are, the better the clustering effect\cite{Ref25}.

\subsubsection{external evaluation measures}
\label{sec:4.1.2}
F-measure(F) is widely used in the fields of statistical classification which combines recall(R) and precision(P) in text clustering\cite{Ref26}. The F-measure controls for the cluster $r$ and category $s$ is decided by the following:

\begin{equation}
P(r,s) = \frac{n_{r,s}}{n_r},
R(r,s)=\frac{n_{r,s}}{n_s}
\end{equation}
where $r$ and $s$ are one of clustering category and one of real class respectively, $n_{r,s}$ is the number of samples both in the real class $s$ and in the cluster $r$. $n_r$ is the number of samples of the cluster $r$, and $n_s$ is the number of samples in class $s$. Then the F value between cluster $r$ and category $s$ is defined as follows.
\begin{equation}
F(r,s)=\frac{2R(r,s)P(r,s)}{R(r,s)+P(r,s)}
\end{equation}
and let $n$ is the number of all texts,$n_r$ means the number of smaples in cluster $r$. F-measure for all clusters is calculated by the following:
\begin{equation}
F=\sum_r \frac{n_r}{n}max \left \{ F(r,s) \right\}
\end{equation}

The Accuracy (Ac) measurement\cite{Ref26} is one of the common external measurements used to compute the percentage of correct assigned documents to each cluster according to the following equation:
\begin{equation}
Ac=\frac{1}{n}\sum_{r=1}^KP(r,s).
\end{equation}
where $P(r,s)$ represent the precision value of real category $s$ in cluster $r$.  K is the number of all clusters.
\subsection{Benchmark evaluation}
\label{sec:4.2}

We use eleven real-word benchmark datasets, and their statistics are summarized in Table 1. The first six datasets \cite{Ref27} are in bag-of-words format after the terms extraction. The first dataset(DS1), called the Computer Science Technical Reports (CSTR)\cite{Ref27}, contain 299 documents which are composed by abstracts and technique reports. CSTR belong to 4 areas: Artificial Intelligence, Robotics, Theory and Systems. SyskillWebert is the second dataset(DS2)\cite{Ref28}, which contain 333 documents and composed by web pages about Goats, Sheep, Biomedical, and Bands. Oh15 is the third dataset (DS3)\cite{Ref29}, is part of the OHSUMED collection\cite{Ref30} and belong to ten topics. The forth, fifth, sixth datasets (DS4-DS6) are Tr11, Tr12, Tr41\cite{Ref29} respectively, they are all from TREC and the topics correspond to the documents that were judged relevant to particular queries\cite{Ref27}.

And the last five datasets are subsets of 20Newsgroups\cite{Ref31}, which was originally collected by Ken Lang. It contains over 18,000 documents in 20 groups. The groups can be divided into 5 categories, where each category has different classes\cite{Ref8}. In order to separate results and reduce computation, we divided the dataset and consider each separately. In the Table, The seventh dataset (DS7), called 20 Newsgroups Computer (20COMP),which contains the group of graphics, ms-windows.misc, pc-hardware, mac.hardware, windows.x that are related to computer technique. The eighth dataset (DS8), called 20 Newsgroups Politics (20POL), contains the group of misc, guns, mideast that are related to politic problems. The ninth dataset (DS9), called 20 Newsgroups Miscellaneous (20MISC), which contains the group of autos, motorcycles, baseball, hockey that are related to sports and vehicle. The tenth dataset (DS10), called 20 Newsgroups Religion (20REL), which contains relision.misc, atheism, christian that are related to religion subject. The eleventh dataset(DS11), called 20 Newsgroups Science (20SCI), which contains the groups of crypt, electronics, med and space that are related to science technique. And these five datasets are in the format of raw documents, so we utilize standard text preprocessing to clean the datasets and extract tokens as introduction in \ref{sec:2.1}. 
\begin{table}[htb]
\begin{threeparttable}
\caption{An overview of the datasets used in our experiments}
\label{1}
  \begin{tabular}{llllllll}
  \hline\noalign{\smallskip}
  Dataset & Abbreviation & Documents & Terms & Tokens & Classes & S-Index\cite{Ref27}  & reference\\ 
  \noalign{\smallskip}\hline\noalign{\smallskip}
  DS1    & CSTR         & 299           & 1725  & 24858  & 4     & 0.752  &\cite{Ref27}  \\
  DS2    & SW           & 333           & 4339  & 58085  & 4    & 0.964   &\cite{Ref28}  \\
  DS3   & Oh15         & 913           & 3100   & 95521  & 10    & 0.857   &\cite{Ref29} \\
  DS4     & Tr11         & 414           & 6429 & 437143  & 9   & 0.937   &\cite{Ref29}  \\
  DS5    & Tr12         & 313           & 5804  & 311111 & 8   & 0.955  &\cite{Ref29}   \\
  DS6 & Tr41         & 878           & 7454     & 357606 & 10  & 0.933  &\cite{Ref29}   \\
  DS7    & 20COMP       & 4582          & 64139 & 420789  & 5   &  0.732 &\cite{Ref8}   \\
  DS8  & 20POL        & 2287          & 41254   & 284113 & 3     &  0.825  &\cite{Ref8}  \\
  DS9  & 20MISC       & 3648          & 39836   & 204655 & 4     &  0.757  &\cite{Ref8}  \\
  DS10   & 20REL        & 2195          & 40149 & 246276 & 3    &  0.889 &\cite{Ref8}   \\
  DS11   & 20SCI        & 3617          & 54373 & 316480 & 4   &   0.831 &\cite{Ref8}  \\ 
  \noalign{\smallskip}\hline
  \end{tabular}
  \begin{tablenotes}
  \item[]Documents only statistics non-empty text number in the dataset. Terms means the number of unique tokens in the dataset. Tokens means the number of all tokens in the dataset. S-index means silhouette coefficient\cite{Ref32} of each classes in each dataset.
  \end{tablenotes}
  \end{threeparttable}
\end{table}

\subsubsection{Parameters setting for the proposed algorithm}
\label{sec:4.2.1}
When we use HMFF technique, parameters need to be set for each document representation measures in Feature Representation Step for different datasets. The evaluation experiments are conducted in order to compare with the baseline and five similar advanced feature selection algorithms. Since these algorithms are used inconsistent parameters which perhaps result in error in the comparative experiments. We strive to ensure the parameter selection under technique is unified to ensure the final results is comparable. In this experiment, All the parameters setting for HMFF and comparing algorithms refers to experts papers as researchers recommend and repeated experiments conducted by ourselves.
\begin{table}[htb]
\begin{threeparttable}
\caption{Parameter setting}
\label{2}
  \begin{tabular}{p{7cm} p{2cm} p{2cm}}
  \hline\noalign{\smallskip}
  Parameter                                 & Symbol & Value      \\
  \noalign{\smallskip}\hline\noalign{\smallskip}
  Random Seed                               & $\sigma$ & 1          \\
  Topic num for topic model                 & $N_{topic}$ & 5$\sim$185 \\
  Iterations for topic model                & $\mu_{topic}$ & 50         \\
  Window size for Word2vec model      & $\theta_{w2v}$ &  1        \\
  Vector dimension for Word2vec model  & $t_{w2v}$ & 200        \\
  Iterations for Word2vec model       & $\mu_{w2v}$ & 100        \\
  Window size for Doc2ec model             & $\theta_{d2v}$ & 5          \\
  Vector dimension for  Doc2Vec model       & $t_{d2v}$ & 200        \\
  Iterations for Doc2Vec model              & $\mu_{d2v}$ & 100        \\
  truncated eigenvectors number for HMFF framework    & $k$ & 3         \\
  Max iterations of k-means algorithm       & $\mu_{kmeans}$ & 300        \\
  \noalign{\smallskip}\hline
  \end{tabular}
    \begin{tablenotes}
    \item[]Since neural network are utilized in distributed word embeddings and both topic model and K-means have random initiate step, we fix Random Seed $\tau=1$ to ensure the repeatability of experimental results. And we set truncated eigenvectors number $k = 3$ in the HMFF framework. Although set different numbers may have a better performance in specific dataset, we still fix this parameter to maintain model consistency.
    \end{tablenotes}
    \end{threeparttable}
\end{table}

Table 2 shows all the parameter we set in the experiment. For the topic model, as the optimal number of topics is related to coherence score in the different data set, we change topic number from 2 to 500 for topic model in each dataset, and calculate each model's coherence score in Fig. 4, and choose the model which has higher score as the optimal model in Table 3.

\begin{figure}[htb]
  \includegraphics[width=1\textwidth]{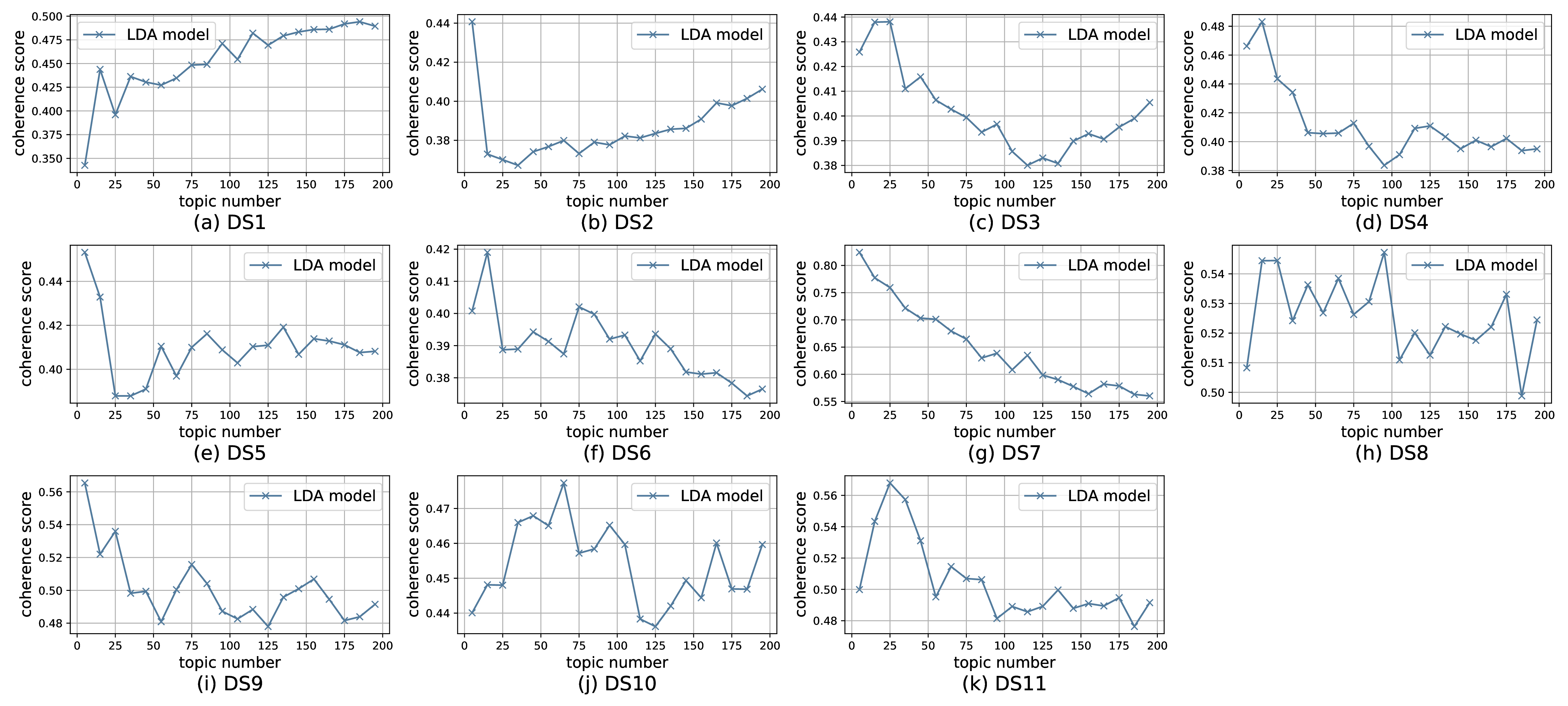}
\caption{coherence score with topic number for different datasets}
\label{fig:4}
\end{figure}
\begin{table}[htb]
\caption{Topic number for different datasets}
\label{3}
  
    \begin{tabular}{l|lllllllllll}
    Dataset & DS1 & DS2 & DS3 & DS4 & DS5 & DS6 & DS7 & DS8 & DS9 & DS10 & DS11 \\ \hline
    $N_{topic}$  & 185  & 5   & 25  & 15  & 5 & 10  & 5 & 95  & 5  & 65   & 25   \\

  \end{tabular}
\end{table}

\subsubsection{Experiment results and discussion}
\label{sec:4.2.2}

Since HMFF framework integrate text representation algorithm from VSM, topic model and distributed Word embeddings (include Word2vec model and Doc2vec model), the first experiment is to examine optimal measurement of each model before conduct integration to let HMFF become effective.

Figure 5 demonstrates the performance of each models with different measurements on these benchmark datasets. To be specific, we evaluated Vector Space models, Topic models and Word2vec models on the datasets (DS1-DS6), and evaluate Doc2Vec models on the datasets (DS7-DS11) as these datasets retain word order. The result shows that applying Cosine-Euclidean measure on VSM and two distributed word embeddings will let models become more effective. And Mallet LDA with Jensen-Shannon measurement performance better among topic models.

\begin{figure}[htb]
  \includegraphics[width=1\textwidth]{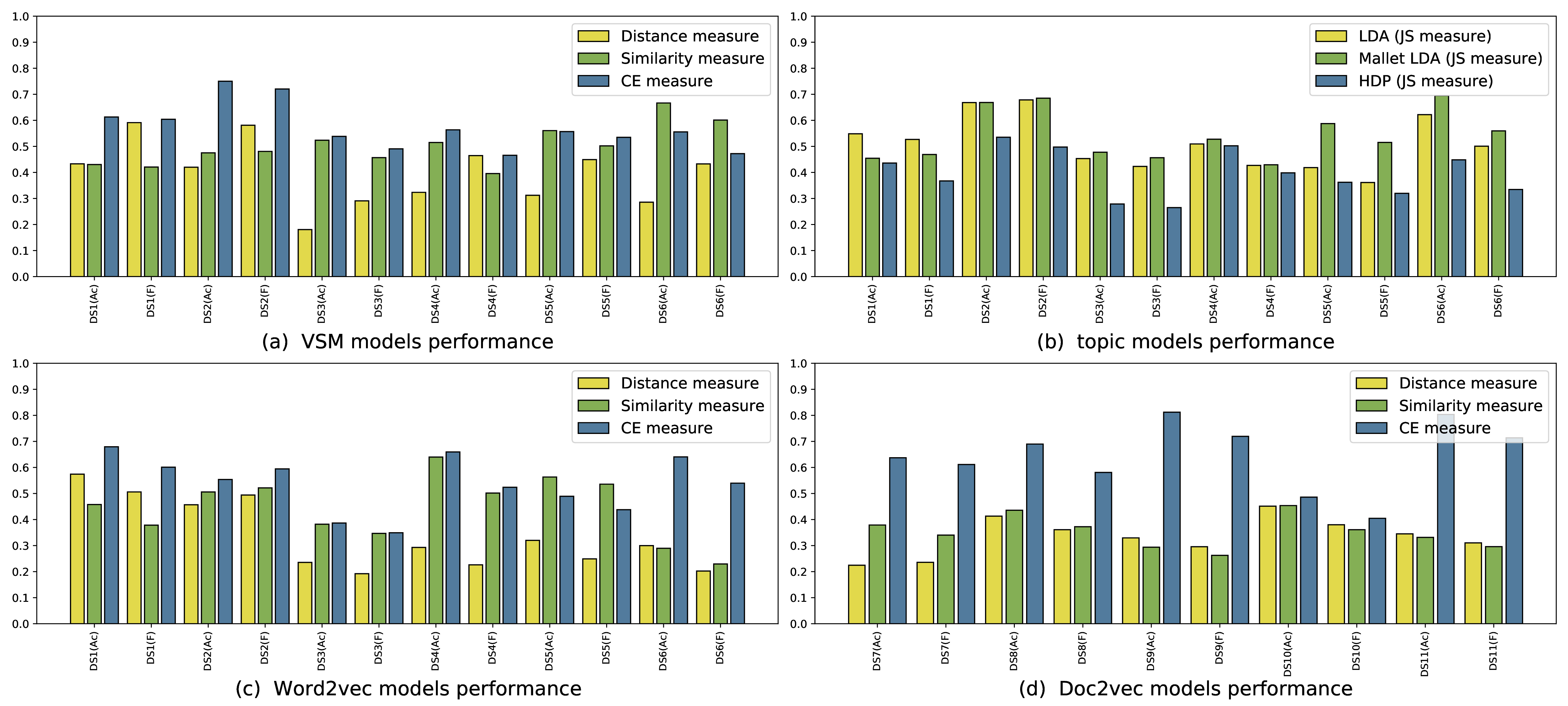}
\caption{Performance of each models with different measurements. Note that the experiment use Distance formula, Similarity formula and Cosine-Euclidean formula (CE measure) to measure VSM [Fig.(a)], Word2vec model [Fig.(c)], Doc2vec model [Fig.(d)] respectively. And compared three topic models of LDA, Mallet LDA\cite{Ref35} and HDP\cite{Ref36}. All topic models used Jensen-Shannon divergence to measure. This experiment use Accuracy (Ac) and F-measure (F) as judgements.} 
\label{fg:5}
\end{figure}

  We conduct comparative evaluations on two different formats of benchmark datasets. And the experiments are carried out to compare HMFF with other four optimization algorithms in each formats. For accurate results and statistical comparisons, we carried out the programs more than 20 times. This number as selected based on the literature which it can sufficient to validate the proposed method\cite{Ref33}. 
  
  DS1-DS6 are bag-of-words format dataset. We choose Word2vec as word embedding model in HMFF word embedding module, and compared this proposed method to Multi-objectives-based K-mean algorithm (MKM), Mallet LDA model (MalletLDA), hybrid PSO algorithm (H-FSPSOTC) and harmony search algorithm (FSHSTC). In the second format, all the datasets are retain word order. We choose Doc2vec as word embedding model in HMFF framework. Comparing methods are MKM, MalletLDA, FSHSTC and Nonnegative Matrix Factorization based text clustering (NMF-FR).

\begin{table}[htb]\scriptsize
\begin{threeparttable}
\caption{Algorithms performance on bag-of-words format}
  \begin{tabular}{ccccccccc}
  \hline\noalign{\smallskip}
  Dataset & Measure    & MKM & MalletLDA & H-FSPSOTC  & FSHSTC & HMFF \\
  ~   &        & \cite{Ref2} & \cite{Ref35} & \cite{Ref9} & \cite{Ref7}  & \\
  \noalign{\smallskip}\hline\noalign{\smallskip}
  DS1     & Accuracy   & 0.6128   & 0.5483   & 0.6792    & 0.6060    & \textbf{0.7640}   \\
  ~       & F-measure  & 0.6041   & 0.5265   & 0.6010    & 0.5808    & \textbf{0.7459}   \\
  ~       & Silhouette & 0.1368   & 0.1261   & 0.0289    & ——        & \textbf{0.2968}   \\
  ~       & Rank       &  2       & 5        &  3        & 4         & \textbf{1}        \\ 
  \hline\noalign{\smallskip}
  DS2     & Accuracy   & 0.7501   & 0.6684   & 0.5538    & 0.7846    & \textbf{0.8560}   \\
  ~       & F-measure  & 0.7203   & 0.6848   & 0.5944    & 0.7537    & \textbf{0.8506}   \\
  ~       & Silhouette & 0.2501   & 0.5160   & 0.0008    & ——        & \textbf{0.2918}   \\
  ~       & Rank       & 3        & 4        & 5         & 2         & \textbf{1}        \\ 
  \hline\noalign{\smallskip}
  DS3     & Accuracy   & 0.5383   & 0.4774   & 0.3966    & \textbf{0.5915}     & 0.5746    \\
  ~       & F-measure  & 0.4906   & 0.4564   & 0.3429    & \textbf{0.5472}     & 0.5456    \\
  ~       & Silhouette & \textbf{0.2199}     & 0.0344    & 0.0246   & ——       & 0.1294    \\
  ~       & Rank       & 3        & 4        & 5         & \textbf{1}          & 2         \\ 
  \hline\noalign{\smallskip}
  DS4     & Accuracy   & 0.5635   & 0.5274   & 0.6594    & 0.6148   & \textbf{0.7721}    \\
  ~       & F-measure  & 0.4659   & 0.4294   & 0.5239    & 0.5311   & \textbf{0.6421}    \\
  ~       & Silhouette & 0.2813   & 0.0476   & 0.0076    & ——       & \textbf{0.3372}    \\
  ~       & Rank       & 4        & 5        & 2         & 2        & \textbf{1}         \\ 
  \hline\noalign{\smallskip}
  DS5     & Accuracy   & 0.5567   & 0.5873   & 0.5632    & 0.6542   &  \textbf{0.7153}   \\
  ~       & F-measure  & 0.5348   & 0.5152   & 0.5358    & 0.6475   & \textbf{0.6766}    \\
  ~       & Silhouette & 0.1956   & 0.0384   & 0.1119    & ——       & \textbf{0.2794}             \\
  ~       & Rank       & 4        & 5        & 3         & 2        & \textbf{1}         \\ 
  \hline\noalign{\smallskip}
  DS6     & Accuracy   & 0.5554   & \textbf{0.6947}      & 0.4110   & 0.5815   & 0.6437    \\
  ~       & F-measure  & 0.4720   & \textbf{0.5594}      & 0.2985   & 0.4536   & 0.5321    \\
  ~       & Silhouette & \textbf{0.3333}     & 0.2009    & 0.0065   & ——       & 0.2365    \\
  ~       & Rank       & 3        & \textbf{1}           & 5        & 3        & 2          \\ 
  \hline\noalign{\smallskip}
  ~ & \textbf{Mean rank}  & 3.17  & 4.00     & 3.83      & 2.33     & 1.33    \\
  ~ & \textbf{Final rank} & 3     & 5        & 4         & 2        & \textbf{1}   \\ 
  \noalign{\smallskip}\hline\noalign{\smallskip}
  \end{tabular}
    \begin{tablenotes}
    \item[]For each dataset, we use Accuracy, F-measure and Silhouette to evaluate cluster's quality, and shows algorithms' comprehensive performance rank based on above three judgement. As H-FSPSOTC and FSHSTC only use Accuracy and F-measure as judgements, so silhouette scores of these two methods in the table are omitted. At last we statistic algorithms' rank in different dataset and give Final rank.
    \end{tablenotes}
    \end{threeparttable}
\end{table}

\begin{table}[htb]\scriptsize
\begin{threeparttable}
\caption{Algorithms performance on token order retained corpus format}
  \begin{tabular}{cccccccc}
  \hline\noalign{\smallskip}
  Dataset & Measure    & MKM & MalletLDA & NMF-FR   & FSHSTC & HMFF \\
  & &\cite{Ref2} &\cite{Ref35} & \cite{Ref8}& \cite{Ref7} &  &  \\
  \noalign{\smallskip}\hline\noalign{\smallskip}
  DS7     & Accuracy   & 0.4131   & 0.3950  & \textbf{0.6370}     & 0.4015    & 0.5476    \\
  ~       & F-measure  & 0.3760   & 0.3586  & \textbf{0.6114}    & 0.3741      & 0.5279    \\
  ~       & Silhouette & 0.0995   & 0.1170  & ——    & ——      & \textbf{0.2154}    \\
  ~       & Rank       & 3        & 5       & \textbf{1}          & 4     & 2    \\ 
  \hline\noalign{\smallskip}
  DS8     & Accuracy   & 0.7158   & 0.4987  & 0.6897     & 0.6945     & \textbf{0.7638}   \\
  ~       & F-measure  & 0.5783   & 0.4190  & 0.5808     & 0.5921     & \textbf{0.6527} \\
  ~       & Silhouette & 0.2084   & 0.0421  & ——      & ——     & \textbf{0.3027}   \\
  ~       & Rank       &   2       & 5       & 4          & 2     & \textbf{1}    \\ 
  \hline\noalign{\smallskip}
  DS9     & Accuracy   & 0.5960   & 0.5408  & \textbf{0.8121}     & 0.6314     & 0.7830    \\
  ~       & F-measure  & 0.5518   & 0.4955  & 0.7194     & 0.5819     & \textbf{0.7342}    \\
  ~       & Silhouette & 0.1492   & 0.1628  & ——      & ——     & \textbf{0.3038}  \\
  ~       & Rank       & 4       & 5       & 2       & 3     &\textbf{1}    \\ 
  \hline\noalign{\smallskip}
  DS10    & Accuracy   & 0.6287   & 0.5206  & 0.4860     & \textbf{0.6700}     & 0.6581   \\
  ~       & F-measure  & 0.5416   & 0.4421  & 0.4045   & \textbf{0.6213}    & 0.5476   \\
  ~       & Silhouette & \textbf{0.2145}   & 0.0285  & ——     & ——     & 0.0655    \\
  ~       & Rank       & 3        & 4       & 5          & \textbf{1}    & 2    \\ 
  \hline\noalign{\smallskip}
  DS11    & Accuracy   & 0.5142   & 0.4759   & 0.8035    & 0.5210    & \textbf{0.8293}    \\
  ~       & F-measure  & 0.4351   & 0.3845   & 0.7137   &  0.4513     & \textbf{0.7516}    \\
  ~       & Silhouette & 0.3098   & 0.0523   & ——   & ——      & \textbf{0.3200}    \\
  ~       & Rank       & 4       & 5        & 2         & 3     & \textbf{1}    \\ 
  \hline\noalign{\smallskip}
  ~ & \textbf{Mean rank}  & 3.20   & 4.80   & 2.80            & 2.60      & 1.40   \\
  ~ & \textbf{Final rank} & 4   & 5   & 3             & 2      & \textbf{1}    \\ 
  \noalign{\smallskip}\hline\noalign{\smallskip}
  \end{tabular}
  \begin{tablenotes}
  \item[] All the judgements are same as which in Table 4. As NMF-FR and FSHSTC only use Accuracy and F-measure as judgements, so silhouette scores of these two methods in the table are omitted. 
  \end{tablenotes}
  \end{threeparttable}
\end{table}
 This section shows the performance of each algorithm based on clusters quality. We can see the results of proposed Hybrid Multisource Feature Fusion (HMFF) framework outperformed all types of basic algorithms shows in Fig.5 based on external measurement. As expected, the HMFF framework improved the text clustering performance when compared with utilizing text represents methods alone. In Table 4, the result shows HMFF have the best performance in DS1, DS2, DS4 and DS5. FSHSTC and MalletLDA win in DS3 and DS5 respectively. And result in Table 5 shows that our proposed method outperforms the other algorithms in DS8, DS9 and DS10, and in the second position in the rest datasets. Generally, comparing algorithms may have better performance in specific dataset but have low rank in another dataset, which means these algorithms become less effective or even malfunctioned in these situations. For example, FSHSTC ranks 1st on DS3 and DS10, but falls to forth place on DS1 and DS7. In comparison, HMFF wins in 7 of 11 benchmark datasets and still keep a high ranking on the rest 4 datasets as well, which performs overall better than other competitors.

\subsection{Wild dataset clustering}
\label{4.3}
  Apart from benchmark tests, we also create a wild real environment as an extended experiment to test HMFF framework performance in practice. In the extended experiment, the HMFF framework was utilized on a twitter dataset about COVID-19 discussion, the dataset consists of 32132 documents and 353668 terms after text preprocessing. We set Doc2vec as word embedding model to build HMFF and set topic number $N_{topic}=137$ for malletLDA, since when $N_{topic}=137$ topic model has the highest coherence score. Then we compared different type of text selection algorithms including HMFF, FSHSTC, NMF-FR, MKM, H-FSPSOTC and MalletLDA. Cluster count change from 4 to 10, which is an explainable range.
  \begin{figure}[htb]
  \centering
    \includegraphics[width=0.75\textwidth]{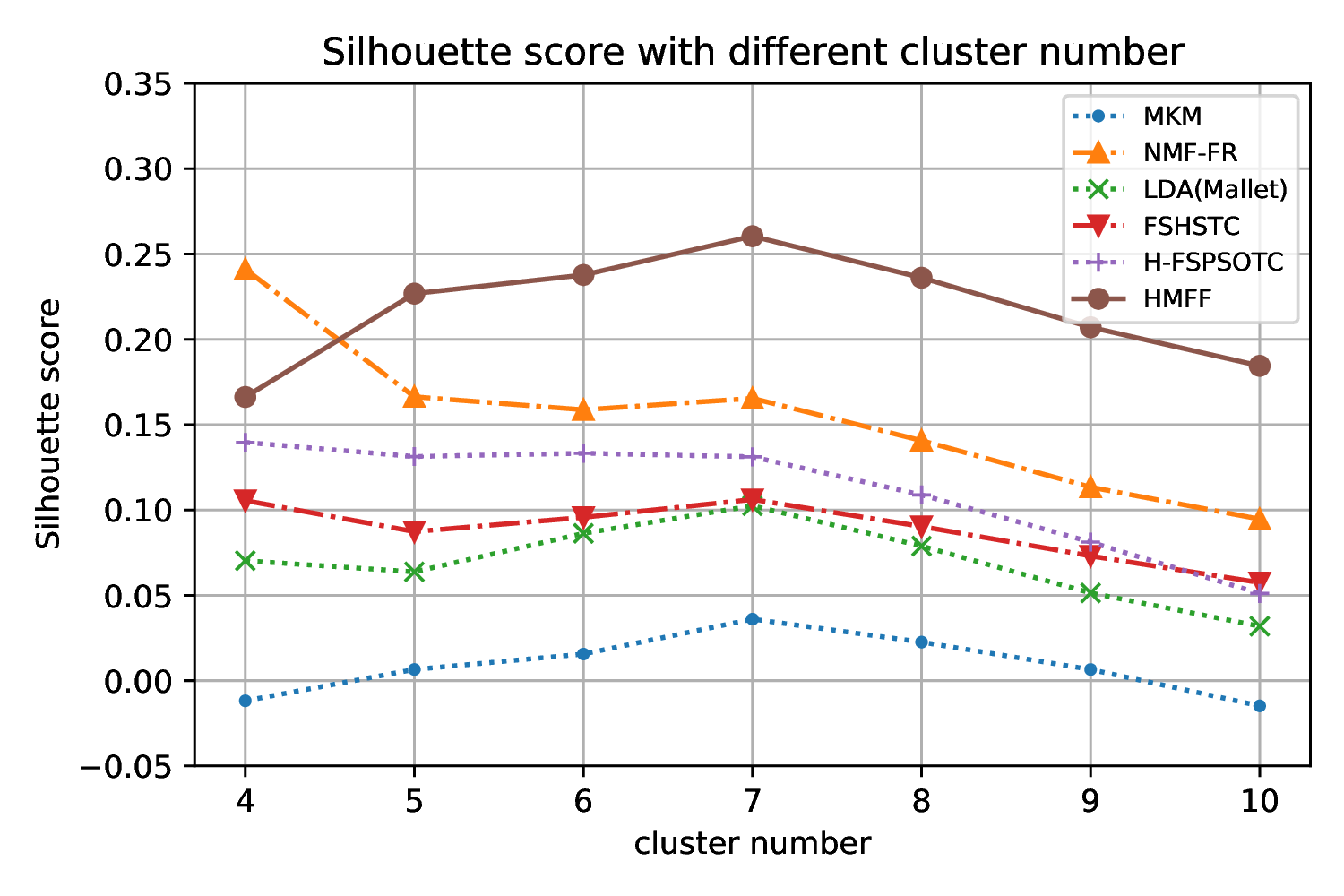}
  \caption{algorithms performance on different cluster number}
  \label{fig:5}
  \end{figure}
  
Although twitter public opinion data is completely stochastic and has no definite categories, which brings huge challenge to the text clustering method, Figure \ref{fig:5} shows HMFF features still could obtain almost the highest silhouette score on each cluster count settings. Furthermore, all these six algorithms unanimously show the most appropriate cluster number of this dataset is 7.


\section{Conclusions and future work}
\label{sec:5}
This paper presents a new feature fusion framework which hybrid three different source of text representation algorithm to extract comprehensive information from documents, namely, Hybrid Multisource Feature Fusion (HMFF) framework. And these HMFF features are used by K-means clustering algorithm to generate accurate clusters. In fact, our HMFF framework could run with other clustering algorithm other than K-means. Six datasets with bag-of-words format and five datasets retaining word order are used for performance and comparing evaluations. All these data sets are public benchmarking text datasets, which are used most widely in comparing algorithms. Five competitors come from recently published papers and compared with HMFF. The experimental results show that the performance of HMFF framework is generally better than other algorithms in each format, which means multisource features of HMFF enhance the accuracy and steadiness of clustering effect. At last, we give a case of applying HMFF framework in the wild corpus, and high adaptability of HMFF framework is shown in the experiment result.

We should note that HMFF framework could accommodate other feature sources from more feature representation algorithms but rather these three sources in section \ref{sec:3.1} only. HMFF framework has high scalable architecture so that feature vectors from additional sources could be easily combined into the HMFF feature matrix without significantly increasing overall feature-computing time. The 3 sources of text representation algorithms in section \ref{sec:3.1} have de facto made superior performance to other competitors. We believe that integrating with more effective feature sources helps to further enhance clustering performance. In our future work, we will keep seeking more effective feature sources to be integrated into HMFF framework.


%
%



\end{document}